\documentclass[10pt,twocolumn,letterpaper]{article}

\usepackage{iccv}              

\usepackage{algorithm,algorithmic}
\usepackage{colortbl}

\usepackage{makecell}

\DeclareMathAlphabet{\mathsfit}{\encodingdefault}{\sfdefault}{m}{sl}
\SetMathAlphabet{\mathsfit}{bold}{\encodingdefault}{\sfdefault}{bx}{n}

\usepackage{pifont}
\newcommand{\cmark}{\textcolor{green}{\ding{51}}}
\newcommand{\xmark}{\textcolor{red}{\ding{55}}}

\definecolor{iccvblue}{rgb}{0.21,0.49,0.74}
\usepackage[pagebackref,breaklinks,colorlinks,allcolors=iccvblue]{hyperref}
\usepackage{bigfoot}
\def\paperID{1140} 
\def\confName{ICCV}
\def\confYear{2025}
\title{\includegraphics[scale=0.3]{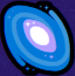} UniVerse: Unleashing the Scene Prior of Video Diffusion Models for\\ Robust Radiance Field Reconstruction}

\author{
    Jin Cao$^{1\dagger}$
    \quad Hongrui Wu$^{2\dagger}$
    \quad Ziyong Feng$^{3}$
    \quad Hujun Bao$^{1}$
    \quad Xiaowei Zhou$^{1}$
    \quad Sida Peng$^{1*}$
    \vspace{1em}
    \\
    $^1$State Key Lab of CAD\&CG, Zhejiang University \quad 
    $^2$Tongji University \quad
    $^3$DeepGlint 
}

\begin{document}
\maketitle

\footnote{$\dagger$ Co-first author. $^*$Corresponding author.}

\begin{abstract}
This paper tackles the challenge of robust reconstruction, i.e., the task of reconstructing a 3D scene from a set of inconsistent multi-view images. 
Some recent works have attempted to simultaneously remove image inconsistencies and perform reconstruction by integrating image degradation modeling into neural 3D scene representations.
However, these methods rely heavily on dense observations for robustly optimizing model parameters.
To address this issue, we propose to decouple robust reconstruction into two subtasks: restoration and reconstruction, which naturally simplifies the optimization process.
To this end, we introduce UniVerse, a unified framework for robust reconstruction based on a video diffusion model. 
Specifically, UniVerse first converts inconsistent images into initial videos, then uses a specially designed video diffusion model to restore them into consistent images, and finally reconstructs the 3D scenes from these restored images.
Compared with case-by-case per-view degradation modeling, the diffusion model learns a general scene prior from large-scale data, making it applicable to diverse image inconsistencies.
Extensive experiments on both synthetic and real-world datasets demonstrate the strong generalization capability and superior performance of our method in robust reconstruction. 
Moreover, UniVerse can control the style of the reconstructed 3D scene. Project page: \href{https://jin-cao-tma.github.io/UniVerse.github.io/}{https://jin-cao-tma.github.io/UniVerse.github.io/} .
\end{abstract}\vspace{-0.3cm}
\section{Introduction}

Novel view synthesis have long been a high-profile and complicated task in computer graphics, which plays a significant role in many applications like virtual reality (VR) and autonomous driving. 
Traditional approaches~\cite{Snavely2008ModelingTW,schonberger2016structure} reconstruct 3D scenes based on the point cloud representation and multi-view stereo techniques.
However, such methods generally suffer from low rendering quality, thus limiting their applications.

In recent years, differentiable rendering-based methods, such as Neural Radiance Fields (NeRF)~\cite{mildenhall2021nerf,barron2021mip,barron2022mip,muller2022instant} and 3D Gaussian Splatting (3DGS)~\cite{kerbl20233d,yan2024street,chen2024pgsr,lee2024c3dgs}, have made significant progress in rendering photorealistic novel views. 
However, these methods assume that all input images are static and captured under consistent conditions. 
In reality, images are frequently affected by illumination variations caused by changes in camera exposure or environmental lighting, content alterations due to dynamic objects, and motion blur resulting from camera shake.
These inconsistencies violate the assumptions of the differentiable-rendering based methods, leading to significant performance degradation~\cite{martin2021nerf,yang2023cross}.


To overcome this problem, previous methods propose learnable embeddings~\cite{martin2021nerf,chen2022hallucinated,tancik2022block,li2023nerf} to additionally model the viewpoint-specific content for each image and jointly optimize it with the underlying 3D scene representation to minimize the rendering loss.
When scene observations are sufficient, these methods can successfully recover the intrinsic scene structure from the inconsistent images.
However, their performance tends to degrades significantly as the number of observations decreases. 
A plausible reason is that they introduce additional learnable parameters into the optimization process, making it more unstable, needing dense observations for optimization.


In this paper, we propose UniVerse, a video generative model for robust 3D reconstruction from inconsistent multi-view images.
Our key idea is to exploit the strong consistent prior of video diffusion models~\cite{yu2024viewcrafter,liu2024reconxreconstructscenesparse} to transform all inconsistent images to a consistent state before performing 3D reconstruction.
Specifically, given a set of unstructured multi-view images, we first sort them to obtain a camera trajectory and insert blank images along this trajectory to transform images into a video.
To better utilize observations, a multiple-input query transformer is proposed to aggregate information from all input images and generate a global semantic embedding, which is injected into the video diffusion model to help the video restoration.
In contrast to previous methods that manually model the degradation in each image, the video diffusion model learns a general consistent scene prior from large-scale data, making it more robust in handling diverse inconsistencies.


We apply UniVerse to both synthetic and real-world challenging inconsistent image collections and demonstrate its ability to produce high-fidelity renderings with fewer artifacts and floaters, surpassing previous state-of-the-art methods in terms of PSNR, SSIM, and LPIPS. By selecting a style image, UniVerse can change the style of the generated videos to match that of the style image, thereby altering the style of the final reconstructed 3D scene. Even when the input images are very sparse and inconsistent (e.g., only 2 images with occlusions), UniVerse can still restore them into convincing consistent images, which can be applied to other downstream tasks such as generating new views~\cite{yu2024viewcrafter} and performing further reconstruction. Overall, these results demonstrate the effectiveness of UniVerse and highlight the potential of decoupling robust reconstruction.

\begin{figure*}
    \centering
    \includegraphics[width=\linewidth]{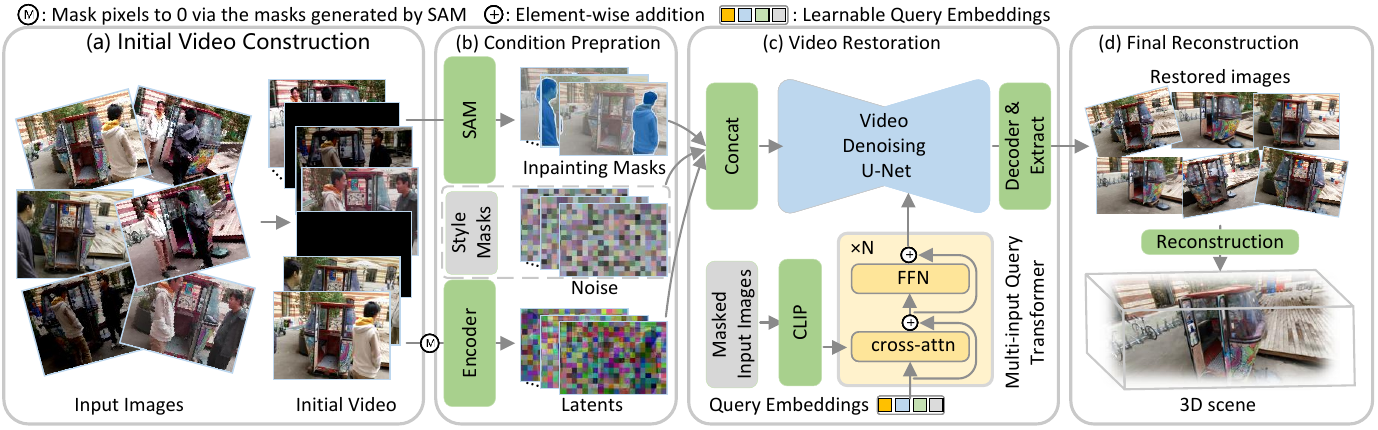}
    \caption{\textbf{The flowchart of UniVerse}. Given a set of inconsistent images, we first convert them into an initial video. We then use SAM~\cite{kirillov2023segany} to identify transient occlusions and generate inpainting masks. These masks are used to set the occluded pixels in the initial video to zero. Next, we encode the video into latents using a VAE Encoder. After setting one image as the style image and assigning it style mask, we concatenate the style masks, inpainting masks, latents, and randomly sampled Gaussian noise along the channel dimension and feed them into the U-Net. For each masked input image, we obtain semantic embeddings using the CLIP image encoder and aggregate them via the Multi-input Query Transformer to form a global semantic embedding. This embedding guides the U-Net in the video generation process. Finally, the U-Net output is decoded by the VAE Decoder to produce the restored video, from which we extract the consistent images and reconstruct a high-quality 3D scene. If too many images for the VDM to restore at once, we iteratively restore them in batches as described. }
    \label{fig:universe}
\end{figure*}

\section{Related Works}

\subsection{Video Diffusion Models for 3D Reconstruction}
The success of diffusion models has also spurred research in diffusion-based video generation~\cite{pku_yuan,blattmann2023svd,chen2023videocrafter1,xing2023dynamicrafter,blattmann2023svd}, which are often fine-tuned from T2I models using extensive video datasets~\cite{bain2021frozen,chen2024panda,xue2022advancing} and can generate consistent videos from text~\cite{pku_yuan,blattmann2023svd,chen2023videocrafter1} or image inputs~\cite{xing2023dynamicrafter,blattmann2023svd}. Recent advancements~\cite{blattmann2023align,chen2023videocrafter1,ge2023preserve,wang2023lavie,zhang2023show,zhou2022magicvideo} further enhance text-to-video generation visual quality through extra temporal layers and curated datasets. The rapid developments of VDMs provoke significant interest in more controllable video generation, enabling controls like RGB images~\cite{blattmann2023svd,xing2023dynamicrafter,xing2024tooncrafter}, depth~\cite{xing2024make,esser2023structure}, trajectory~\cite{yin2023dragnuwa,niu2024mofa}, and semantic maps~\cite{peruzzo2024vase}. Recently, some works further explore camera motion control for VDMs to generate controllable 3D-aware videos~\cite{guo2023animatediff,blattmann2023svd,wang2023motionctrl,muller2024multidiff}. Recently, CamCo~\cite{xu2024camco} and CameraCtrl~\cite{he2024cameractrl} introduced Plücker coordinates~\cite{sitzmann2021light} in video diffusion models for camera motion control. ViewCrafter~\cite{yu2024viewcrafter} and ReconX~\cite{liu2024reconxreconstructscenesparse} further use explicit point clouds to achieve more precise 3D-aware camera control. These works achieve great success in generating consistent 3D-aware videos which can be directly used to reconstruct a 3D scene. Observing the strong consistent 3D prior of VDMs and noting that multi-view images are similar to frames extracted from a video captured by a camera trajectory, we take inconsistent multi-view images as conditions and generate a consistent video, and then extract them from the generated video with all images being consistent and static.

\subsection{Robust 3D Reconstruction}

Reconstructing a 3D scene from a set of 2D images is a long-standing problem in computer vision. Modern approaches, such as NeRF-based methods~\cite{muller2022instant,fridovich2022plenoxels,yu2021plenoctrees,garbin2021fastnerf, reiser2021kilonerf} and 3DGS-based methods~\cite{kerbl20233d,lu2023scaffold,fan2023lightgaussian,yu2023mip}, have achieved great success and demonstrated expressive reconstruction quality. However, these methods all assume that the input images are captured in a static scene. Their performance declines significantly when reconstructing from unconstrained inconsistent photo collections. To address this challenging in-the-wild task, several attempts~\cite{meshry2019neural,martin2021nerf,rudnev2022nerf,chen2022hallucinated,yang2023cross,li2023nerf,kulhanek2024wildgaussians} have been made to handle appearance variation and transient occlusions. Other works~\cite{li2020crowdsampling,lin2023neural} focus on scenes with time-varying appearances, while methods~\cite{zhang2021ners,kuang2022neroic,engelhardt2024shinobi} use physical rendering models for diverse lighting conditions. Nevertheless, these methods typically couple the process of restoration and reconstruction and directly perform reconstruction on the inconsistent multi-view images, which requires a large amount of training images to identify and remove all inconsistencies. What's more, they typically use handcraft prior to manually model inconsistency in each image~\cite{wang2024bilateral}. In contrast, our method emphasizes the effectiveness of restoration before reconstruction, decoupling the task and making it much easier. Meanwhile, the VDM we use learns a general consistent scene prior from large-scale data, thus can be more robust facing various inconsistencies. Recently, SimVS~\cite{trevithick2024simvs} utilized a multi-view diffusion model~\cite{gao2024cat3d} to turn all images into a consistent state given an image as a reference. However, it retains all inconsistencies of the reference image, such as a moving passenger, thus can fail to reconstruct the static scene,  while our method removes all inconsistencies in the images and aims to reconstruct the static scene.

\section{Method} \label{sec:method}


 \paragraph{Background.} 3D reconstruction aims to recover the 3D structure of a scene from multiple 2D images taken from different viewpoints. While traditional methods like structure-from-motion~\cite{Snavely2008ModelingTW,schonberger2016structure} have been widely used,  newer techniques such as NeRF~\cite{quei2020nerf} and 3DGS~\cite{kerbl20233d} leverage differentiable rendering. Given input images $\{I_i\}_{i=1}^K$ and their corresponding poses $\{P_i\}_{i=1}^K$, differentiable rendering aims to find a parameterized function $f_\theta$ that takes a camera pose as input and outputs the corresponding image. The goal is to optimize the parameters $\theta$ or the 3D representation to minimize the following loss function:
\begin{equation}
    \theta = \mathop{\arg\min}_{\theta} \sum_{i=1}^{K} Dif(f_\theta(P_i), I_i). \label{eq:diff_render}
\end{equation}
Here, $Dif(\cdot)$ is a differentiable function, such as MSE or L1 loss, used to measure the difference between two images. Once $\theta$ is obtained, we can render novel views from the 3D scene for any new camera pose $P$ using $f_\theta(P)$, thereby achieving 3D reconstruction. However, these approaches assume that the the images $\{I_i\}_{i=1}^K$  are consistent and static. If the assumption isn't hold, the learning of $f_\theta$ fails. To address this, UniVerse uses a VDM to restore all input images to be static and consistent. This ensures that the assumption for Eq.~(\ref{eq:diff_render}) is valid, helping $f_\theta$ to easilly learn the 3D scene.

 \paragraph{Overview.} Given $K$ inconsistent multi-view images $\{I_i\}_{i=1}^K, I_i \in \mathbb{R}^{3 \times H \times W}$, our goal is to restore them into $K$ consistent images. 
We treat the multi-view images as video frames from the same video. 
We first sort the images into ordered $K$ images based on their camera poses , as described in in Sec.~\ref{sec:thread}. Then we use a VDM to restore all these $K$ images into a consistent state. Assuming the VDM can generate $f$ frames at a time, we iteratively restore all images, processing $N \leq f$ images per iteration. 
We select one of the first $N$ images as the style image $I_{sty}$, and turn the first $N$ images into a initial video as described in Sec. \ref{sec:method:turn_i2v}. For all frames in the initial video, we assign them inpainting masks to indicate where to be inpainted and style masks to indicate which frame should be taken as the style reference via Segment Anything Model (SAM)~\cite{kirillov2023segany}. Using the initial video, inpainting masks, and style masks as conditions, we use the VDM to generate a restored video  with the same style as the style image, extracting the corresponding $N$ consistent frames. We then remove the corresponding $N$ images from the input unrestored $K$ images since they are already restored, and add the last restored image to the unrestored images  as the first image, and set it as style image for the next iteration. 
We update $K$ to $\max(K - (N - 1), 1)$ and repeat the process until $K \leq 1$, which means all images are restored. Finally, with all restored $K$ images, we use 3D reconstruction methods like NeRFs~\cite{quei2020nerf} to reconstruct them and return a 3D representation. We show this process in Fig.\ref{fig:universe} and Alg. \ref{alg:universe} in \textbf{Supp}.

\subsection{Turning Multi-view Images into Videos} \label{sec:method:turn_i2v}

As discussed, UniVerse first converts the $K$ input multi-view images into initial videos. These images are essentially captured by cameras at various poses around a single scene. Assuming all $K$ poses $\{P_i\}_{i=1}^K$ lie on a single camera trajectory, continuously sampling new poses (and thus new views) from this trajectory yields a video if the poses are sufficiently dense. This approach hinges on solving two key problems: (I) determining the trajectory and (II) sampling new poses from it. The detailed algorithm is provided in Sec. \ref{appendix:more_details}, Alg. \ref{alg:threadpose} and Fig. \ref{fig:turn_i2v} the \textbf{Supp}.


 \paragraph{Sorting Multi-view Images for Sparse Trajectory.}\label{sec:thread} We use ThreadPose to sort the $K$ input poses $\{P_i\}_{i=1}^K$ into an ordered list. We initialize a double linked list with a randomly chosen pose and iteratively add the remaining poses based on their distances to the current head and tail of the list. The distance metric combines rotation and translation differences, weighted to ensure consistent scaling. Finally, we traverse the whole list to construct an ordered set of poses $\{P'_i\}_{i=1}^K$, which defines an implicit camera trajectory.


 \paragraph{Sample Implicit Dense Views from Trajectory} \label{sec:method:sample_views} At each iteration, given $N$ ordered poses $\{P'_i\}_{i=1}^N$ and the corresponding $N$ inconsistent images $\{I'_i\}_{i=1}^N$, our goal is to create an initial video of $f$ frames that includes all $N$ input images. We achieve this by sampling $f - N$ new poses from the trajectory to generate new views. First, we compute the distances $\{d_i\}_{i=1}^{N-1}$ between neighboring poses:
\begin{equation}
    d_i = D_P(P'_i, P'_{i+1}), \quad i = 1, 2, \ldots, N-1.
\end{equation}
Here, $D_P(\cdot)$ is a function to compute the distance between two poses, which is defined in \textbf{Supp}. Next, we assign the number of new poses $n_i$ to be inserted between each pair of neighboring poses $P'_i$ and $P'_{i+1}$, proportional to the distance $d_i$. After assigning the number of poses, we aim to sample new poses and new views. However, since the input images are inconsistent, it can be difficult to use methods like conditional VDM generation~\cite{he2024cameractrl,xu2024camco}, or building 3D structures like point clouds~\cite{yu2024viewcrafter}, to explicitly render a new view given an explicit camera pose. Considering VDM's strong 3D prior and frame interpolation ability, we simply insert $n_i$ zero frames into each $I'_i, I'_{i+1}$ neighboring image pair and expect the VDM to inpaint these zero frames into new views. In this way, we turn $N$ ordered images $\{I'_i\}_{i=1}^N$ into an $f$-frame initial video with the first image $I'_1$ and the last image $I'_N$ as the first and last frames.

\subsection{Conditional VDM for Initial Video Restoration} \label{sec:method:VDM}


 \paragraph{Preliminary: Video Diffusion Models.} \label{subsec:prelimiary} In diffusion-based video generation, Latent Diffusion Models (LDMs)~\cite{metzer2022latent} are often employed to reduce computational costs. In LDMs, video data ${x} \in \mathbb{R}^{f \times 3 \times H \times W}$ is encoded into the latent space using a pre-trained VAE encoder frame-by-frame, expressed as ${z} = \mathcal{E}({x})$, ${z} \in \mathbb{R}^{f \times C \times h \times w}$. The forward and reverse processes are then performed in the latent space. The final generated videos are obtained through the VAE decoder $\hat{{x}} = \mathcal{D}({z})$. In this work, we build our VDM based on an open-sourced Image-to-Video (I2V) diffusion model DynamiCrafter~\cite{xing2023dynamicrafter}.

 \paragraph{Initial Video Restoration.} As shown in Fig.~\ref{fig:universe}, at a certain iteration, given the input images and the corresponding initial video, inpainting masks, and style masks, we first use the VAE encoder to encode the initial video into latents and downsample both masks to match the shape of the latents. We then concatenate the latents, inpainting masks, style masks, and randomly sampled Gaussian noise along the channel dimension to form the image inputs.

To better leverage the I2V VDM's ability to control video generation via text-aligned semantic embeddings~\cite{xing2023dynamicrafter}, we extend the Query Transformer in~\cite{xing2023dynamicrafter} to a Multiple-input Query Transformer. Considering we have $N$ input images per iteration, we pass them through the CLIP image encoder~\cite{radford2021clip} to obtain $N$ embeddings, which are then injected into the Multiple-input Query Transformer via cross-attention as the value and key. This yields a global semantic embedding to aid in generating 3D-aware videos.

With the image inputs and the embedding as conditional inputs, we use the VDM to generate restored latents and decode them using the VAE Decoder to produce the consistent restored video. Finally, we extract the corresponding $N$ frames from the restored $f$-frame video as the restored images. We discard the other $f-N$ new views, as they can be unreliable without 3D-consistency constraints.


 \paragraph{Training VDMs for Consistency.} As discussed, UniVerse aims to make input images consistent. The purpose of sampling dense views, as described in Sec.~\ref{sec:method:sample_views}, is to transform discrete images into a video to leverage the VDM's prior, rather than generating new views. Directly training the VDM using MSE Loss is inappropriate because the simple MSE loss equally weights making the input $N$ frames consistent and generating $f - N$ new views. Given $2 \leq N < f$, we need to adjust the loss weights for each of the $f$ frames to ensure the VDM focuses on making existing images consistent. To this end, we propose a consistency loss $\mathcal{L}_{con}$ for VDM training. Specifically, assuming the initial video is $\mathbf{v}^{ini} \in \mathbb{R}^{f \times 3 \times H \times W}$, given the VDM's estimated noise $\epsilon_\theta$ and the ground truth noise $\epsilon \in \mathbb{R}^{f \times C \times h \times w}$ during training, we first compute the MSE loss frame-by-frame to obtain the loss vector $lv \in \mathbb{R}^f$:

\begin{equation}
    lv[i] = \| \epsilon_\theta[i] - \epsilon[i] \|_2^2, \quad \text{for } i = 1, 2, \ldots, f,
\end{equation}
where $[i]$ denotes the $i$-th frame. We then adjust $lv$ as follows:
\begin{equation}
    lv[i] = lv[i] \times \begin{cases} 
\omega_{c} & \text{if } \mathbf{v}^{ini}[i] \text{ is an input image}, \\
\omega_n & \text{otherwise (i.e., } \mathbf{v}^{ro}[i] \text{ is a zero frame)}.
\end{cases}
\end{equation}
The weights $\omega_c$ and $\omega_n$ are computed as:
\begin{equation}
    \omega_c = \max\left(\frac{N}{f}, \lambda\right) \big/ \frac{N}{f},
\end{equation}
\begin{equation}
    \omega_n = \min\left(\frac{f - N}{N}, 1 - \lambda\right) \big/ \frac{f - N}{f}.
\end{equation}
This ensures that the ratio of the loss weights for making images consistent to generating new views is at least $\lambda : (1 - \lambda)$. In practice, we set $\lambda$ to a large value like 0.98. Additionally, since our VDM takes initial video, inpainting/style masks as input, we need a special training data construction approach, which we discuss in Sec.~\ref{sec:experiment:imple}.

\begin{figure*}[t]
    \centering
    \includegraphics[width=\linewidth]{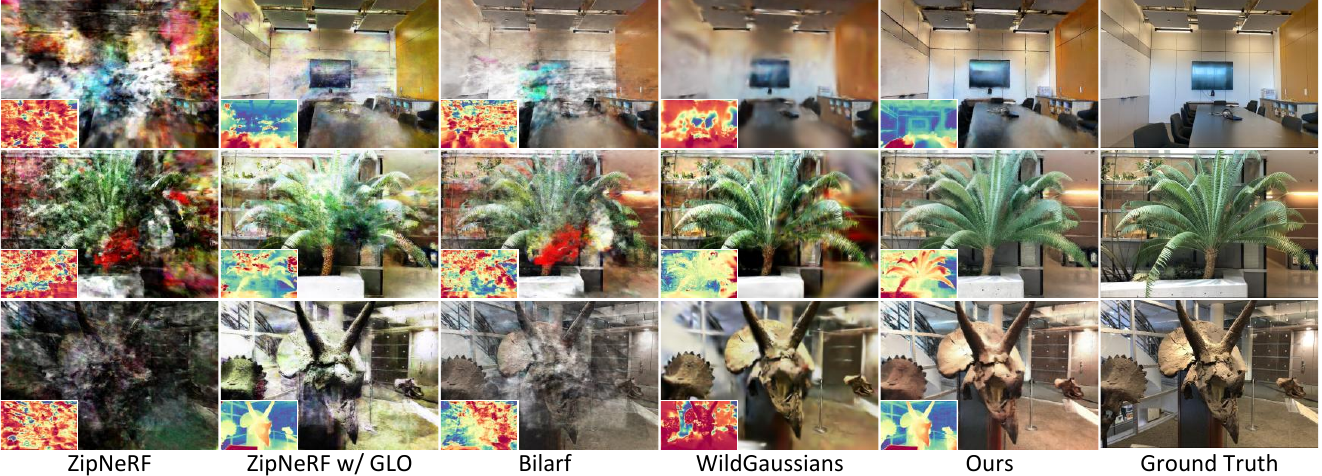}
    \caption{Visual results of novel view synthesis on synthetic datasets, with the corresponding depth map displayed in the bottom left corner.}
    \label{fig:syn_visual}
\end{figure*}

\begin{figure*}[t]
    \centering
    \includegraphics[width=\linewidth]{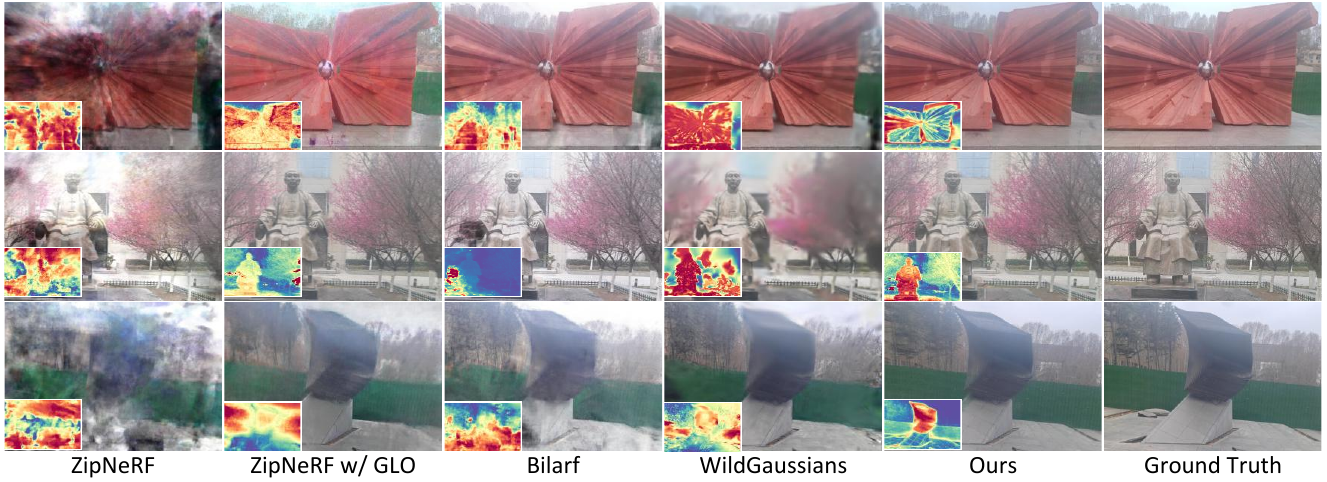}
    \caption{Visual results of novel view synthesis on real datasets, with the corresponding depth map displayed in the bottom left corner.}
    \label{fig:real_visual}
\end{figure*}

\section{Experiment}
\subsection{Implementation Details} \label{sec:experiment:imple}

\par \textbf{VDM Training Details:} We employ a progressive training strategy to fine-tune the VDM. Specifically, we fine-tune the $576 \times 1024$ interpolation VDM from ViewCrafter~\cite{yu2024viewcrafter}. In the first stage, we train the VDM at a resolution of $320 \times 512$, with the frame length $f$ set to 25. The entire video denoising U-Net is fine-tuned for 14,520 iterations using a learning rate of $5 \times 10^{-5}$ and a batch size of 8. In the second stage, we continue to fine-tune the video denoising U-Net at a resolution of $576 \times 1024$ for high-resolution adaptation, with 12,000 iterations on a learning rate of $1 \times 10^{-5}$ and a mini-batch size of 8. All training is conducted on 8 NVIDIA A100 GPUs.
\par \textbf{Training Data Construction:} Our VDM was trained on the DL3DV dataset~\cite{ling2024dl3dv}. Specifically, we first extract 25 frames from the video of DL3DV at a random FPS to simulate the varying levels of density and sparsity in real-world multi-view images. Then we randomly set $n, 0 \leq n \leq 23$ of the 25 frames to zeros. Next, for each frame in the 25 frames, we randomly adjust their brightness, sharpness, hue, saturation, and simultaneously add Gaussian noise, motion blur, Gaussian blur, and occlusions to simulate inconsistencies. We use the VOC2007 dataset~\cite{pascal-voc-2007} to generate the occlusions and inpainting masks. In the masks, "1" indicates that this pixel needs to be inpainted, while "0" indicates the opposite. For zero frames, the inpainting masks are filled with "1". In this way, we generate a initial video and corresponding inpainting masks. Finally, we randomly choose a non-zero frame from the initial video as the style image and generate the style masks. Specifically, for the style frame, the mask is all "1", while for other frames, the mask is "0". We then adjust all frames in the original video to match the style of the style image to obtain the target video. In this way, we generate a training pair. In total, we generate 116,158 video pairs as training data.
\par \textbf{Inferencing Details:} During inference, we adopt the DDIM sampler~\cite{song2021denoising} with classifier-free guidance~\cite{ho2022classifier}. We use SegNeXt~\cite{segnext} as the semantic segmentation model to identify transient occlusions in the input images, and then use SAM~\cite{kirillov2023segany} to segment these occlusions and generate the inpainting masks. Assuming $O$ is the minimum integer such that $\frac{K-1}{O} < 25$, we set the number of images processed at each iteration to $N=\lfloor \frac{K-1}{O} \rfloor +1$. After all images are made consistent, we use ZipNeRF~\cite{zipnerf} with GLO~\cite{martin2021nerf} to reconstruct the 3D scene. All inference is conducted on a single NVIDIA A100 GPU.

\begin{figure}
    \centering
    \includegraphics[width=\linewidth]{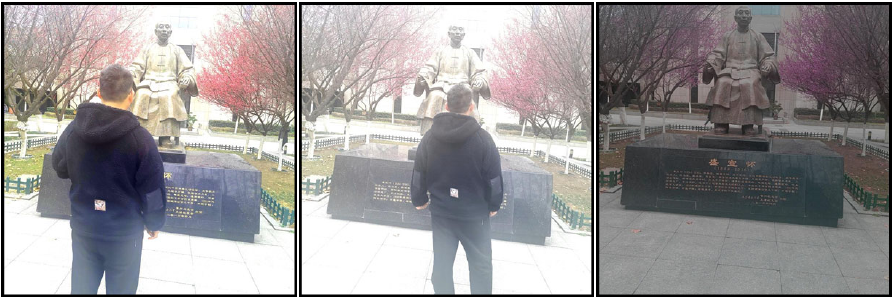}
    \caption{Samples of our captured images. }
    \label{fig:capture_samples}
\end{figure}

\subsection{Evaluation}
We aim to evaluate the ability of UniVerse to alleviate the problem mentioned in the \textbf{Background} of Sec.~\ref{sec:method}, \ie the problem of inconsistent images. We evaluate our method on both synthetic datasets and real-world datasets and compare it with the latest methods.
\par
\textbf{Dataset:} For synthetic datasets, we utilize the NeRF llff dataset~\cite{quei2020nerf}. Since all images in this dataset are captured under static, consistent conditions, we randomly adjust their brightness, sharpness, hue, and saturation. We also randomly add Gaussian noise, motion blur, Gaussian blur, and occlusions to simulate inconsistencies. To stress test all methods, we limit the number of images for a single scene to 20-50. For real-world datasets, we use cell phone cameras to collect 7 real-world scenes for evaluation. During capture, in addition to the automatic adjustments by camera programs, we manually change local exposure and ISO settings, and apply random post-processing filters to each view. We show samples of our captured images in Fig.~\ref{fig:capture_samples}.
\par 
\textbf{Metrics and Compared Methods:} For quantitative comparison, we use PSNR, SSIM~\cite{wang2004image}, and LPIPS~\cite{zhang2018unreasonable} as metrics to assess the performance of our method. Meanwhile, we calculate a per-channel affine transformation to align the output color tints with the ground truth tints (Affine-aligned sRGB)~\cite{wang2024bilateral}. We also present rendered images generated from the same pose as the input view for visual inspection. To demonstrate the superiority of our method, we compare it against the following methods: ZipNeRF~\cite{zipnerf}, ZipNeRF W/GLO~\cite{martin2021nerf}, Bilarf~\cite{wang2024bilateral}, and WildGaussians~\cite{kulhanek2024wildgaussians}.
\par 

\begin{table}[tb]
    \centering
    \caption{The quantitative results in novel view synthesis on synthetic datasets. The best and second-best results are highlighted.}
    \resizebox{\linewidth}{!}{
    \begin{tabular}{l|c|c|c|c|c|c}
        \toprule
        \textbf{} & \multicolumn{3}{c|}{sRGB} & \multicolumn{3}{c}{Affine-aligned sRGB} \\
        \textbf{} & PSNR $\uparrow$  & SSIM $\uparrow$ & LPIPS $\downarrow$ & PSNR $\uparrow$ & SSIM $\uparrow$ & LPIPS $\downarrow$ \\ \midrule
        ZipNeRF & 11.52 & 0.2327 & 0.6930 & 14.53 & 0.3548 & 0.6297 \\
        ZipNeRF w/GLO & 14.53 &0.4667 &  \cellcolor{yellow!30}0.4394 & \cellcolor{yellow!30}18.58 & \cellcolor{yellow!30}0.5297 & \cellcolor{yellow!30}0.4123 \\
        Bilarf & \cellcolor{yellow!30}16.06 & \cellcolor{yellow!30}0.4814 & 0.4489 & 17.68 & 0.5062 & 0.4317 \\ 
        WildGaussians & 13.75 & 0.3972 & 0.6430 & 15.45 & 0.4476 & 0.6244 \\ 
        Ours & \cellcolor{red!30}18.09 & \cellcolor{red!30}0.5789 & \cellcolor{red!30}0.3015 & \cellcolor{red!30} 20.12& \cellcolor{red!30} 0.5926& \cellcolor{red!30} 0.2979 \\ \bottomrule
    \end{tabular}
    }
    \label{tab:quanti_eval}
\end{table}

\begin{table}[tb]
    \centering
    \caption{The quantitative results in novel view synthesis on real-world datasets. The best and second-best results are highlighted.}
    \resizebox{\linewidth}{!}{
    \begin{tabular}{l|c|c|c|c|c|c}
        \toprule
        \textbf{} & \multicolumn{3}{c|}{sRGB} & \multicolumn{3}{c}{Affine-aligned sRGB} \\
        \textbf{} & PSNR $\uparrow$  & SSIM $\uparrow$ & LPIPS $\downarrow$ & PSNR $\uparrow$ & SSIM $\uparrow$ & LPIPS $\downarrow$ \\ \midrule
        ZipNeRF & 13.31 & 0.5222 & 0.4310 & 16.38 & 0.5490 & 0.4328 \\
        ZipNeRF w/GLO & \cellcolor{yellow!30}16.14 &\cellcolor{yellow!30}0.5668 &  \cellcolor{yellow!30}0.3095 & \cellcolor{yellow!30}20.67 & 0.5907 & \cellcolor{yellow!30}0.3125 \\
        Bilarf & 15.46 & 0.5459 & 0.3323 & 18.84 & \cellcolor{yellow!30}0.6055 & 0.3169 \\ 
        WildGaussians & 15.05 & 0.4246 & 0.5046 & 17.39 & 0.5747 & 0.4493 \\ 
        Ours & \cellcolor{red!30}19.65 & \cellcolor{red!30}0.6511 & \cellcolor{red!30}0.2532 & \cellcolor{red!30} 22.91& \cellcolor{red!30} 0.6998& \cellcolor{red!30} 0.2132 \\ \bottomrule
    \end{tabular}
    }
    \label{tab:quanti_eval_real}
\end{table}

\textbf{Results:} We present the average quantitative results in Tab.~\ref{tab:quanti_eval} and Tab.~\ref{tab:quanti_eval_real}, and the qualitative visual results in Fig.~\ref{fig:syn_visual} and Fig.~\ref{fig:real_visual}. Both tables show that UniVerse achieves the best metric values under both settings. Moreover, the figures demonstrate that our method provides the most visually pleasing results with fewer artifacts and floaters. In contrast, other compared methods often produce novel views with noticeable artifacts and a significant number of floaters due to unstable optimization and the lack of dense observations, thereby highlighting the superiority of our approach.

\subsection{Abalation Study}

\textbf{The Effect of the Design of Turning Images into Videos:} As discussed, transforming multi-view images into videos is crucial for unleashing the consistent 3D prior of VDMs. To validate this, we tested the following settings for image restoration and reconstruction: \textit{(a)} Directly stacking unordered images as the initial video. \textit{(b)} Sorting images using ThreadPose and stacking them as the initial video. \textit{(c)} Inserting zero frames (implicit views) into unordered images to form the initial video. \textit{(d)} Inserting zero frames into ordered images, as described in Sec. \ref{sec:method:turn_i2v}. The results in Tab. \ref{tab:abl:i2v} show that only our design fully exploits the 3D prior of VDMs, demonstrating its rationality.

\begin{table}[h]
    \centering
    \small
    \caption{Results on novel view synthesis with different image to video settings.}
        \resizebox{\linewidth}{!}{\begin{tabular}{lcccc}
            \toprule
                Setting &  \textit{ThreadPose} & \textit{zero frames} & PSNR & SSIM \\
             \midrule

            \textit{(a) (directly stack)}  & \xmark & \xmark  & 15.32 & 0.4598 \\
            \textit{(b) (w/ ThreadPose)} &  \cmark & \xmark  &  17.29	& 0.5876 \\
            \textit{(c) (w/ zero frames)}  & \xmark & \cmark  &  18.25	& 0.6067\\
             \textit{(d) (ours)}  & \cmark & \cmark & {\cellcolor{red!30}{20.71}} & {\cellcolor{red!30}{0.7708}}\\
             \bottomrule \\
            \end{tabular}}
    
    \label{tab:abl:i2v}
\end{table}

\begin{figure*}
    \centering
    \includegraphics[width=\linewidth]{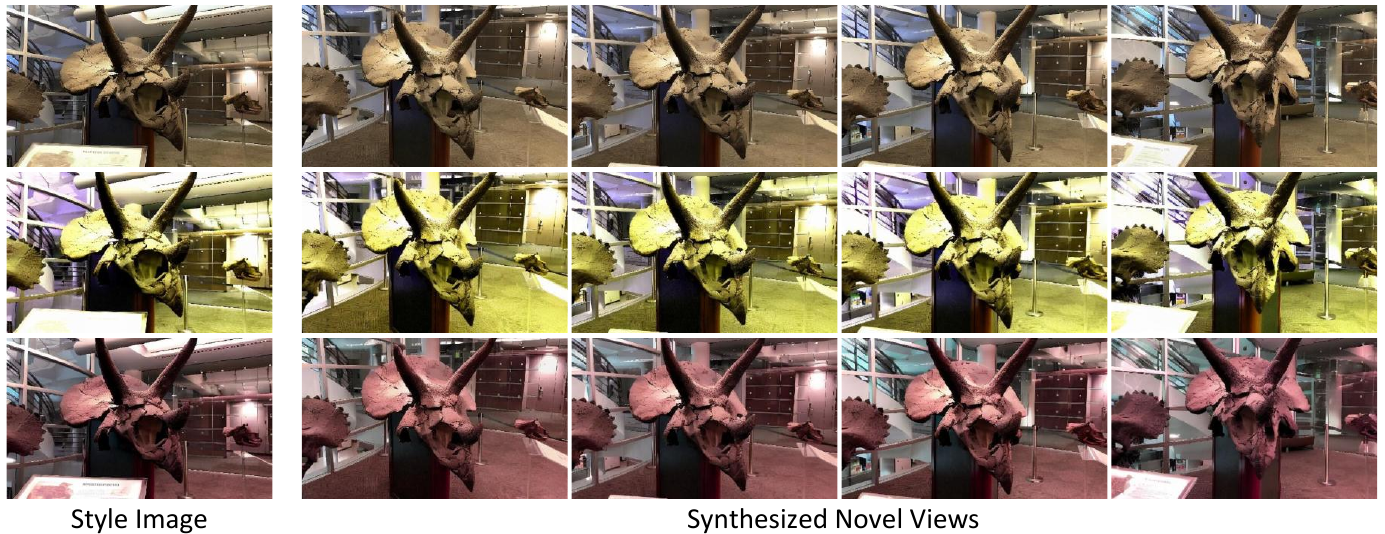}
    \caption{Controlling the style of reconstructed 3D scene by swiching style image.}
    \label{fig:control_style}
\end{figure*}

\begin{figure*}
    \centering
    \includegraphics[width=\linewidth]{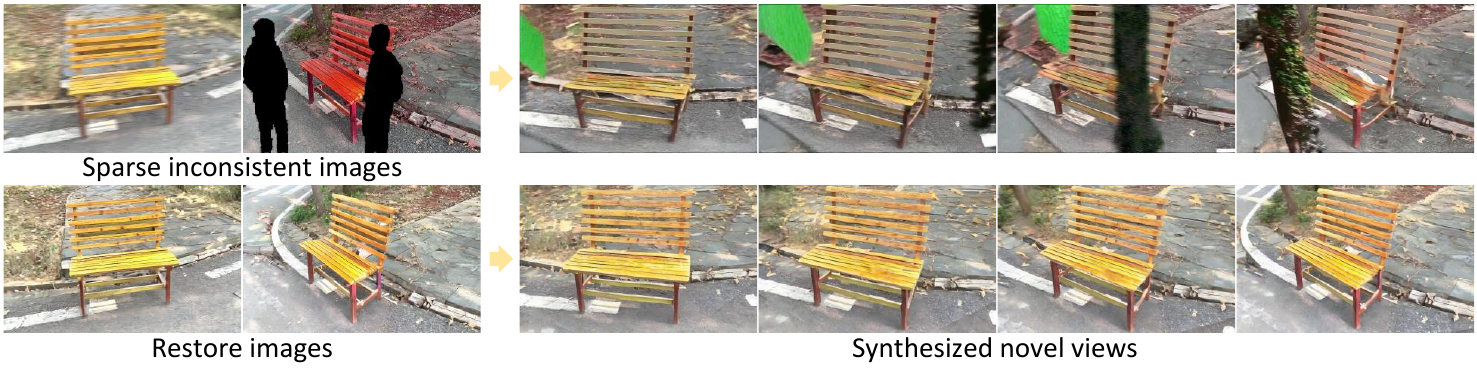}
    \caption{Novel views synthesized via ViewCrafter~\cite{yu2024viewcrafter}. \textbf{First line}: ViewCrafter synthesizes inconsistent and distorted views from inconsistent images. \textbf{Second line:} After the images are restored by UniVerse, the generated views become consistent.}
    \label{fig:sparse}
\end{figure*}

\textbf{The Effect of our VDM Design:} We introduce four new designs for VDMs in this work: the Multi-input Query Transformer (MiQT), style mask, inpainting mask, and consistent loss. We validate each design as follows:
\par 
\textbf{(1) Multi-input Query Transformer (MiQT):} To assess MiQT's impact, we retrained a model using a single-input QT and compared its robust reconstruction performance with ours. The results in Tab. \ref{tab:abl:QT} show MiQT's superiority over QT. This highlights the importance of global semantic information in leveraging VDMs' prior, thereby justifying our design.
\begin{table}[h]
    \centering
    \scriptsize
    \caption{Results on novel view synthesis with different Query Transformer (QT) settings.}
        \resizebox{\linewidth}{!}{\begin{tabular}{lccc|lccc}
            \toprule
                Setting & PSNR & SSIM & LPIPS & Setting & PSNR & SSIM & LPIPS\\
             \midrule
            QT & 16.80 & 0.4157  & 0.5015 & MiQT & \cellcolor{red!30} 17.42 &\cellcolor{red!30} 0.4511  &  \cellcolor{red!30}0.4549\\
             \bottomrule \\
            \end{tabular}}
    \label{tab:abl:QT}
\end{table}
\par
\begin{figure}[h]
    \centering
    \includegraphics[width=\linewidth]{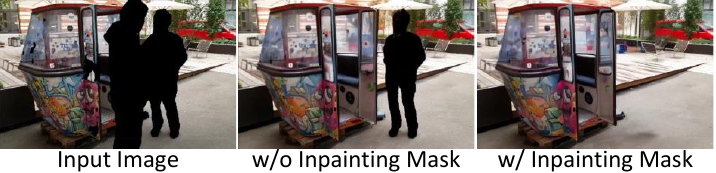}
    \caption{Visualization of results w/o and w/ inpainting masks.}
    \label{fig:abl_inp_mask}
\end{figure}

 
\textbf{(2) Inpainting Mask:} We retrained a VDM without inpainting masks, forcing it to decide which pixels to inpaint. Using a subset of the NeRF-on-the-go dataset~\cite{Ren2024NeRF} containing only occlusions as inconsistencies, we found that the VDM failed to inpaint all masked pixels (Fig. \ref{fig:abl_inp_mask}). Thus, inpainting masks are essential for UniVerse.\par 
\begin{figure}[h]
    \centering
    \includegraphics[width=\linewidth]{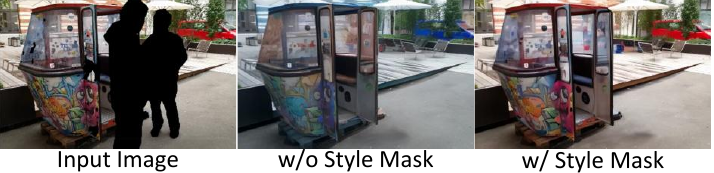}
    \caption{Visualization of results w/o and w/ style masks.}
    \label{fig:abl_sty_mask}
\end{figure}

\textbf{(3) Style Mask:} Figure~\ref{fig:abl_sty_mask} compares results with and without the style mask. Without it, output image tone is uncontrollable despite consistent input tones. Style masks are thus crucial for controlling image appearance and reconstructed 3D scene style.

\par 
\textbf{(4) Consistent Loss:} As elaborated in Sec.~\ref{sec:method:VDM}, employing the Consistent Loss is pivotal for directing the VDM to prioritize image consistency. The comparative training outcomes presented in Tab.~\ref{tab:abl:loss} underscore the superior performance of the Consistent Loss over the regular MSE. This superiority indicates that the Consistent Loss effectively enables the VDM to more adeptly eliminate inconsistencies, thereby substantiating the efficacy of our design approach.

\begin{table}[h]
    \centering
    \scriptsize
    \caption{Results on novel view synthesis with different training loss function settings.}
        \resizebox{\linewidth}{!}{\begin{tabular}{lccc|lccc}
            \toprule
                Setting & PSNR & SSIM & LPIPS & Setting & PSNR & SSIM & LPIPS\\
             \midrule
            MSE & 16.19 & 0.3788  & 0.5340 & Consistent & \cellcolor{red!30} 17.42 &\cellcolor{red!30} 0.4511  &  \cellcolor{red!30}0.4549\\
             \bottomrule \\
            \end{tabular}}
    \label{tab:abl:loss}
\end{table}

\subsection{Further Applications of UniVerse}

\textbf{Controlling the Style of Reconstructed 3D Scene:} The style of the restored images, and consequently the reconstructed 3D scenes, is determined by the style image. By changing the style image, we can alter the style of the entire reconstructed 3D scene, as shown in Fig.~\ref{fig:control_style}.

\textbf{Robust Reconstruction on Sparse Images:} UniVerse focuses on making images consistent rather than generating new views. Thus, even after restoring very sparse input images to a consistent state, reconstruction may still fail due to insufficient views. This issue can be easily resolved by using a generative novel view synthesis model like ViewCrafter~\cite{yu2024viewcrafter}. As shown in Fig.~\ref{fig:sparse}, given 2 inconsistent input images, ViewCrafter~\cite{yu2024viewcrafter} synthesizes distorted novel views with strange occlusions. After the images are restored via UniVerse, the novel views synthesized by ViewCrafter become consistent. In other words, as a restoration model, UniVerse can serve as a pre-processor for other models, enabling robust reconstruction.

\section{Conclusion \& Limitation}
This paper proposes UniVerse, a unified robust reconstruction framework that converts inconsistent multi-view images into initial videos and leverages Video Diffusion Models to restore them into consistent images. By decoupling robust reconstruction into two subtasks (\ie restoration and reconstruction), UniVerse overcomes the limitations of existing approaches that require very dense observations to reconstruct inconsistent images, achieving state-of-the-art performance on both synthetic and real-world datasets. Moreover, we explore UniVerse's ability to control the style of the reconstructed 3D scene by switching the reference image and its potential for reconstructing very sparse inconsistent observations by applying novel view generation models after restoration. We believe our work offers new insights of decoupling robust reconstruction and restoring images using models with 3D priors to the community.

\textbf{Limitations:} UniVerse requires synthesizing videos with inconsistencies as training data to fine-tune the VDM for adaptation to a restoration model. While some inconsistencies, like lighting, may be hard to synthesize, \cite{trevithick2024simvs} have shown tremendous promise for inconsistency synthesis via generative models.

\section{Acknowledgment}

This work was partially supported by the National Key R\&D Program of China (No. 2024YFB2809102), NSFC (No. 62402427, NO. U24B20154), Zhejiang Provincial Natural Science Foundation of China (No. LR25F020003), DeepGlint, Zhejiang University Education Foundation Qizhen Scholar Foundation, and Information Technology Center and State Key Lab of CAD\&CG, Zhejiang University.


{
    \small
    \bibliographystyle{ieeenat_fullname}
    \bibliography{main}
}

\clearpage

\setcounter{page}{1}
\maketitlesupplementary

\begin{figure*}[!h]
    \centering
    \includegraphics[width=\linewidth]{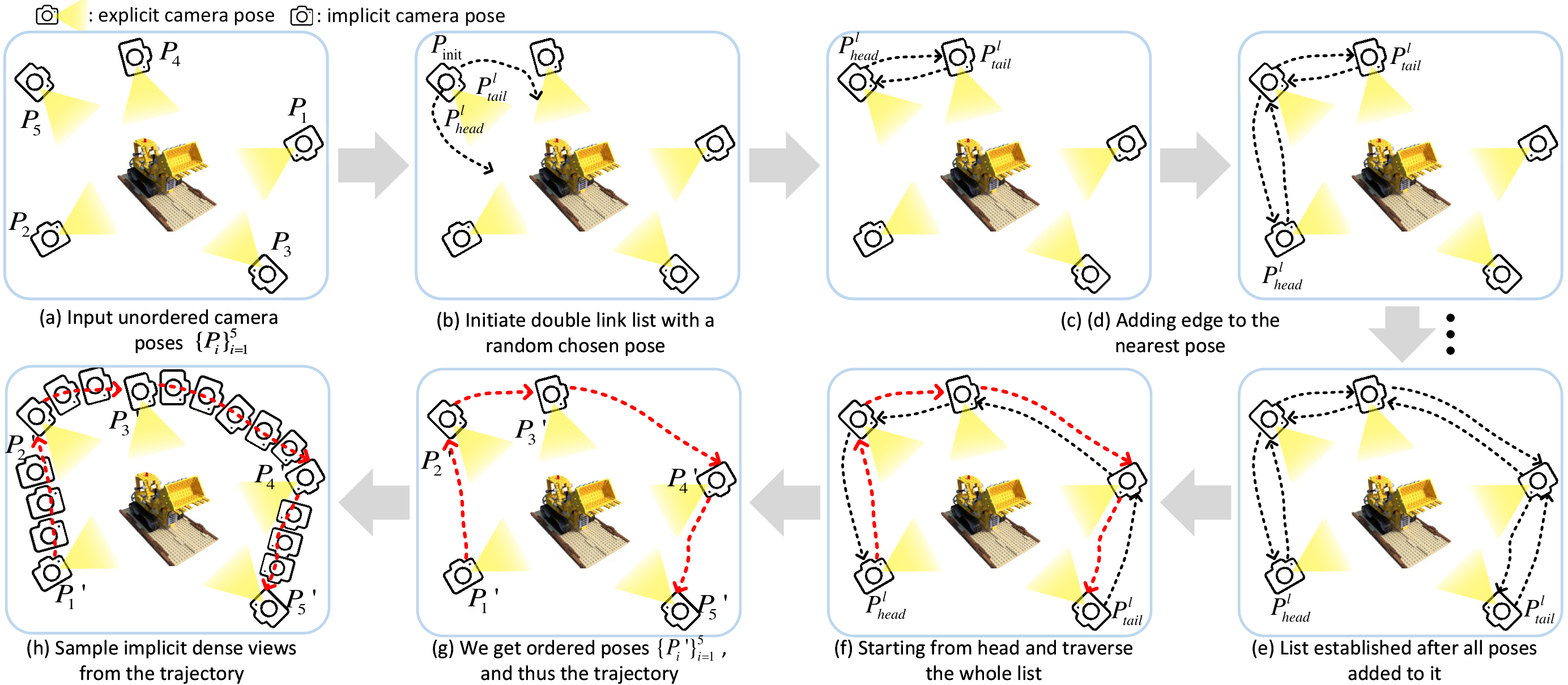}
    \caption{The flowchart of how we transform a set of multi-view images into a initial video. Here we take an example with 5 input images and their poses. Given 5 unordered poses shown in (a), we firstly random choose a pose to initiate a double link list in (b). Next, we iteratively add the nearest pose to the list until all poses are in the list, shown in (c)(d)(e). Then in (f) we start from the head of the list and traverse the whole list and obtain the ordered poses in (g). Finally we add new poses to the intervals of input poses, making the trajectory dense and thus transform images to video.}
    \label{fig:turn_i2v}
\end{figure*}

\section{More Details for Method} \label{appendix:more_details}
\subsection{Sorting Images for Sparse Trajectory.}
Starting with $K$ poses $\{P_i\}_{i=1}^K$, we initialize a double linked list $\{P^{l}_i\}_{i=1}^L$ with a randomly chosen pose $P_{\text{init}} \in \{P_i\}_{i=1}^K$, where $L$ is the length of the list. 
At each iteration, for any pose $P_c \in \{P_i\}_{i=1}^K \setminus \{P^l_i\}_{i=1}^L$, we calculate its distance with the list $D_{PL}$ as follows:
\begin{equation}
    D_{PL}(P_c,\{P^l_i\}_{i=1}^L)=\min\{D_P(P_c,P^l_{head}),D_P(P_c,P^l_{tail})\}.
\end{equation} 
Here, $P^l_{head}$ and $P^l_{tail}$ are the head and tail of the current list, i.e., $P^l_1$ and $P^l_L$. The distance between poses $D_P$ is defined as:
\begin{equation}
    D_P(P_a,P_b)= \frac{\omega_r}{s_R} \cdot D_R(R_a,R_b) + \frac{1-\omega_r}{s_T} \cdot D_T(T_a,T_b).
\end{equation}
Here, $R_a$ and $T_a$ are the rotation matrix and translation vector of the pose $P_a$, respectively. $\omega_r$ is the weight for rotation distance, and $s_R$ and $s_T$ are scale factors to ensure rotation and translation distances have the same scale. We calculate the rotation distance $D_R$ as:
\begin{equation}
    D_R(R_a,R_b) = \arccos\left(\frac{\text{trace}(R_aR_b)-1}{2}\right),
\end{equation}
and the translation distance $D_T$ as:
\begin{equation}
    D_T(T_a,T_b)=\|T_a-T_b\|_2.
\end{equation}
After calculating the distances of all $P_c$ and $\{P^l_i\}_{i=1}^L$, we add the new pose $P_{new}$ with minimal distance to the list:
\begin{equation}
    P_{new}=\mathop{\arg\min}_{P_{c} \in \{P_i\}_{i=1}^K \setminus \{P^l_i\}_{i=1}^L} D_{PL}(P_{new},\{P^l_i\}_{i=1}^L).
\end{equation}
If $P_{new}$ is closer to $P^l_{head}$, we add an edge from $P^l_{head}$ to $P_{new}$ and turn $P_{new}$ into $P^l_{head}$; otherwise, we do the same for $P^l_{tail}$. We iteratively perform this process until all poses in $\{P_i\}_{i=1}^K$ are added to the list. After that, we start from $P_{head}$ and traverse the whole list by edges to get an ordered set of poses $\{P'_i\}_{i=1}^K$ (i.e., $\{P^l_i\}_{i=1}^K$). 

According to $\{P'_i\}_{i=1}^K$, we obtain the ordered images $\{I'_i\}_{i=1}^K$. Along the ordered poses $P'_1, P'_2, \ldots, P'_K$, we actually obtain an appropriate implicit camera trajectory. We show this process in both Alg.~\ref{alg:threadpose} and Fig. \ref{fig:turn_i2v}.

\begin{algorithm}[!h]
\caption{UniVerse}
\label{alg:universe}
\textbf{Input:} Inconsistent multi-view images $\{I_i\}_{i=1}^K$, rough camera poses $\{P_i\}_{i=1}^K$, camera pose estimation method $Camera(\cdot)$ conditional video diffusion model $\mathcal{V}(\cdot)$, number of images per iteration $N$, pose sort function $ThreadPose(\cdot)$, the function to turn images to initial videos $I2V(\cdot)$, transient occlusions segment model $Seg(\cdot)$, 3D Reconstruction Method $Recon(\cdot)$

\begin{algorithmic}[1]
\STATE {\bfseries Initialization:} $I^{re}{'} \leftarrow \{\}$ \\
$\{I'_i\}_{i=1}^K, \{P'_i\}_{i=1}^K$ $\leftarrow$ $ThreadPose ( \{I_i\}_{i=1}^K , \{P_i\}_{i=1}^K ) $  \\
$I_{sty} \leftarrow \text{manually/random choose an image from $\{I'_i\}_{i=1}^N$}$

\WHILE{$K >1 $}
    \STATE Initiate inpainting and style masks $M^{in}\leftarrow\{\},M^{st}\leftarrow \{\}$
    \STATE Extract the first $N$ images: $\{I'_i\}_{i=1}^N$
    \STATE  $\mathbf{v}^{ini} \leftarrow I2V(\{I'_i\}_{i=1}^N, f)$, $\mathbf{v}^{ini}$ refers to initial video
    \FOR{each frame $v_j$ in $\mathbf{v}^{ini}$}
        \IF{$v_j \in \{I'_i\}_{i=1}^N$}
            \STATE Mask transient occlusions: $M^{in}_j, v_j \leftarrow Seg(v_j)$ 
        \ELSE
            \STATE Fill the inpainting mask $M^{in}_j$ with "1"
        \ENDIF
        \IF{$v_j \textbf{ is } I_{sty}, $}
            \STATE Fill style mask $M^{st}_j$  with "1".
        \ELSE
            \STATE Fill style mask $M^{st}_j$  with "0".
        \ENDIF

    \STATE $M^{in}.append(M^{in}_j),M^{st}.append(M^{st}_j)$
    \ENDFOR
    \STATE $\mathbf{v}^{re} \leftarrow \mathcal{V}(\mathbf{v}^{ini},M^{in},M^{st})$
    \STATE Extract the restored images $\{I_i^{re}{'}\}_{i=1}^N$ from $\mathbf{v}^{re}$
    \STATE $I^{re}{'} \leftarrow I^{re}{'}\cup \{I_i^{re}{'}\}_{i=1}^N$ 
    \STATE $I_{sty} \leftarrow I_N^{re}{'}$
    \STATE $\{I'_i\}_{i=1}^K \leftarrow \{I'_i\}_{i=N+1}^K, K \leftarrow \max(K - N, 0)$ 
    \STATE  $\{I'_i\}_{i=1}^K \leftarrow \{I_N^{re}{'}\} \cup \{I'_i\}_{i=1}^{K}$ 
    \STATE Update $K \leftarrow K+1$
\ENDWHILE 
\STATE \# now we get consistent images $I^{re}{'}$ (\ie $\{I_i^{re}{'}\}_{i=1}^K$)
\STATE $\{P'_i\}_{i=1}^K \leftarrow Camera(I^{re}{'})$ \# estimate poses again using consistent images 
\STATE {\bfseries Output: the reconstructed 3D scene $Recon(I^{re}{'},\{P'_i\}_{i=1}^K)$  } 

\end{algorithmic}
\end{algorithm}

\begin{algorithm}[!h]
\caption{ThreadPose for Implicit Camera Trajectory}
\label{alg:threadpose}
\textbf{Input:} Poses $\{P_i\}_{i=1}^K$, $add\_edge(\cdot)$ func to add bidirectional edges, $Traverse(\cdot)$ func to traverse the list by edges
\begin{algorithmic}[1]
\STATE Initialize a double linked list $\{P^{l}_i\}_{i=1}^L$ with a randomly chosen pose $P_{\text{init}} \in \{P_i\}_{i=1}^K$
\STATE Set $L \leftarrow 1$, $P^{l}_1 \leftarrow P_{\text{init}}$
\WHILE{$ \{P_i\}_{i=1}^K \setminus \{P^l_i\}_{i=1}^L$ is \textbf{not empty}}
    \STATE Find the pose $P_{new}$ with the minimal distance:
    \begin{equation*}
        P_{new} = \mathop{\arg\min}_{P_{c} \in \{P_i\}_{i=1}^K \setminus \{P^l_i\}_{i=1}^L} D_{PL}(P_{new}, \{P^l_i\}_{i=1}^L)
    \end{equation*}
    \IF{$D_P(P_{new}, P^l_{head}) < D_P(P_{new}, P^l_{tail})$}
        \STATE $P_{head}.add\_edge(P_{new})$
        \STATE $P^{l}_{head} \leftarrow P_{new}$
    \ELSE
        \STATE $P_{tail}.add\_edge(P_{new})$
        \STATE $P^{l}_{tail} \leftarrow P_{new}$
    \ENDIF
    \STATE $L \leftarrow L + 1$
\ENDWHILE
\STATE $\{P'_i\}_{i=1}^K \leftarrow Traverse(P_{head})$

\STATE {\bfseries Output:} Ordered poses $\{P'_i\}_{i=1}^K$
\end{algorithmic}
\end{algorithm}

\subsection{Sampling Implicit Views}

At each iteration, given $N$ ordered poses $\{P'_i\}_{i=1}^N$ and corresponding $N$ inconsistent images $\{I'_i\}_{i=1}^N$, our goal now is to create a initial video of $f$ frames inluding all the input $N$ images. And we inflate it to $f$ frames by sampling $f-N$ new poses and thus new views. First, we compute the distances $\{d_i\}_{i=1}^{N-1}$ between neighboring poses:
\begin{equation}
    d_i = D_P(P'_i, P'_{i+1}), \quad i = 1, 2, \ldots, N-1.
\end{equation}

Next, we determine the number of new poses $n_i$ to be inserted between each pair of neighboring poses $P'_i$ and $P'_{i+1}$, proportional to the distance $d_i$:
\begin{equation}
n_i = \left\lfloor \frac{d_i}{\sum_{i=1}^{N-1} d_i} \times (f - N) \right\rfloor, \quad i = 1, 2, \ldots, N-1,
\end{equation}
where $\left\lfloor x \right\rfloor$ denotes the floor function, which gives the greatest integer $\leq x$. Since the sum of $n_i$ might not exactly equal $f - N$ due to the floor operation, we distribute the remaining poses. We calculate the remaining number of poses $r$:
\begin{equation}
r = (f - N) - \sum_{i=1}^{N-1} n_i.
\end{equation}
Then, we add one additional pose to the $r$ largest intervals (i.e., the intervals with the largest $d_i$ values) by incrementing $n_i$ for the $r$ largest $d_i$ values:
\begin{equation}
n_i = n_i + \begin{cases} 
1 & \text{if } d_i \text{ is among the } r \text{ largest values}, \\
0 & \text{otherwise}.
\end{cases}
\end{equation}

In this way, we obtain the number of inserted views. By inserting $n_i$ zero frames into neighboring images $I_i',I'_{i+1}$, we get the initial video.

\section{More Implementation Details}
\paragraph{Adapt Video Diffusion Models with Mask Input:} We fine-tune the Video Diffusion Model from the $576 \times 1024$ interpolation model of ViewCrafter~\cite{yu2024viewcrafter}. Since our method utilizes additional masks (\ie inpainting masks and style masks), we need to change the input dimension of the Denoising U-Net. We follow the fine-tuning approach of Inpainting Latent Diffusion~\cite{metzer2022latent}. Specifically, we change an $8 \times C \times \textit{kernel\_size} \times \textit{kernel\_size}$ 2D convolutional kernel to $10 \times C \times \textit{kernel\_size} \times \textit{kernel\_size}$ by concatenating two additional masks. To do this, we maintain the original $8 \times C \times \textit{kernel\_size} \times \textit{kernel\_size}$ kernels and add zero-initialized $2 \times C \times \textit{kernel\_size} \times \textit{kernel\_size}$ kernels to it.

\paragraph{Detect All Transient Objects in Input Images:} In the UniVerse pipeline, it is important to identify all transient objects to mask them. To achieve this, we first pre-define a set of transient prompts, such as \texttt{[person, car, bike]}. We then use a Semantic Segmentation Model to detect the pixels of the objects in the prompts. Using the positions of these pixels, we employ the Segment Anything Model (SAM)~\cite{kirillov2023segany} to precisely segment the objects and obtain the inpainting masks.
\section{More Visual Results}
\begin{figure*}
    \centering
    \includegraphics[width=0.95\linewidth]{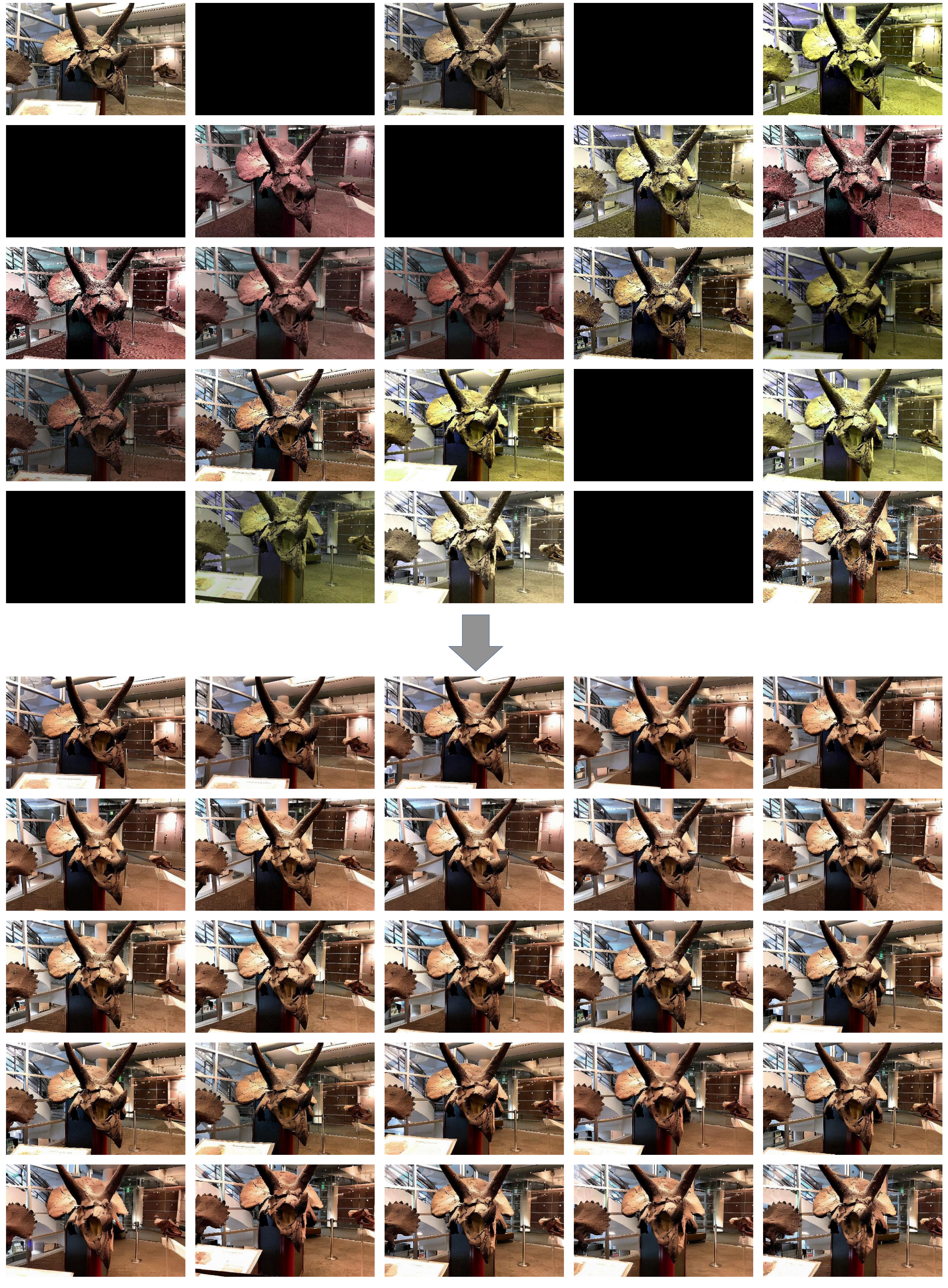}
    \caption{Visualization of how UniVerse turns a initial video into restored video.}
    \label{fig:app_video_horn}
\end{figure*}

\begin{figure*}
    \centering
    \includegraphics[width=0.95\linewidth]{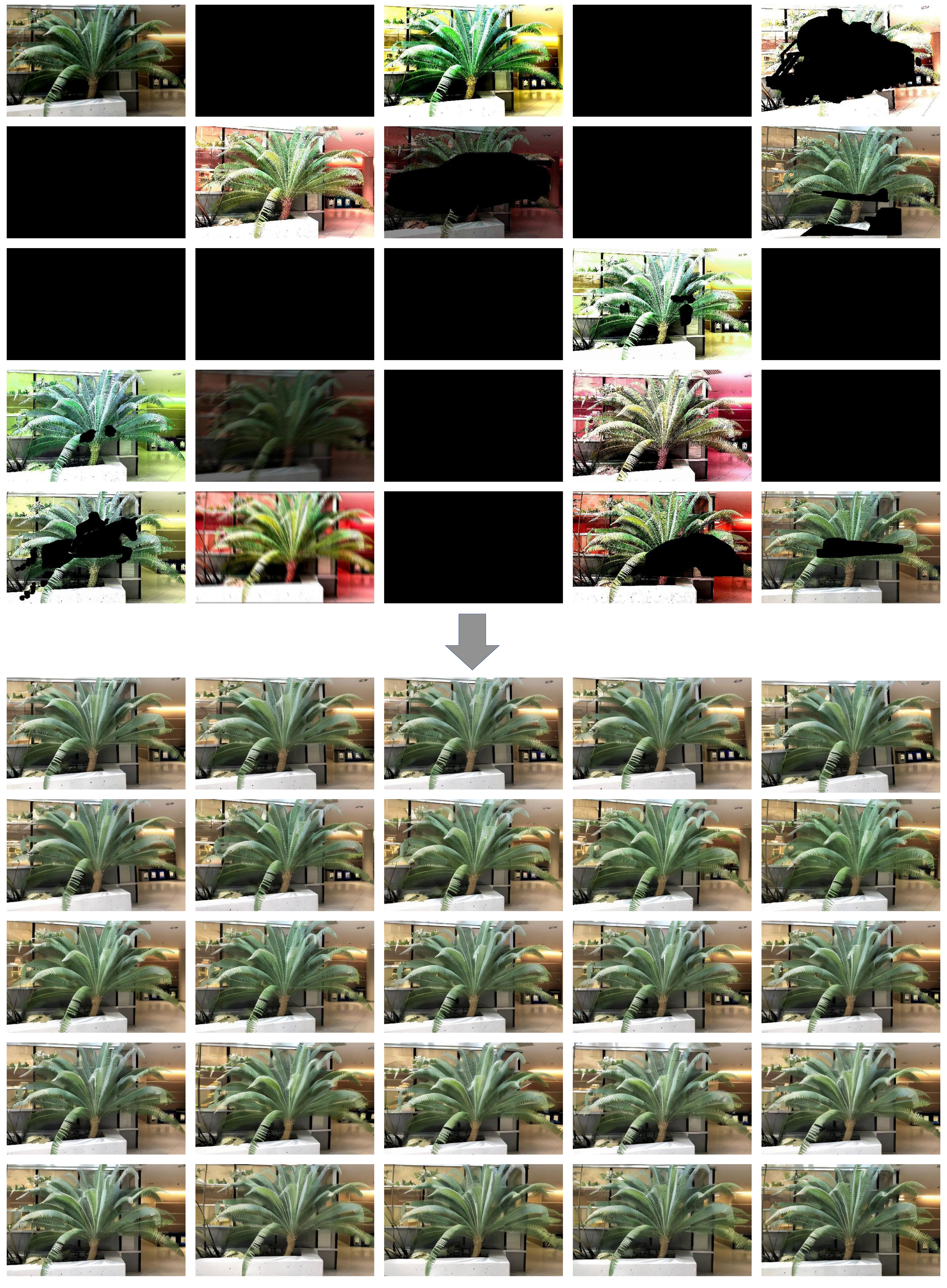}
    \caption{Visualization of how UniVerse turns a initial video into restored video.}
    \label{fig:app_video_fern}
\end{figure*}

\begin{figure*}
    \centering
    \includegraphics[width=0.95\linewidth]{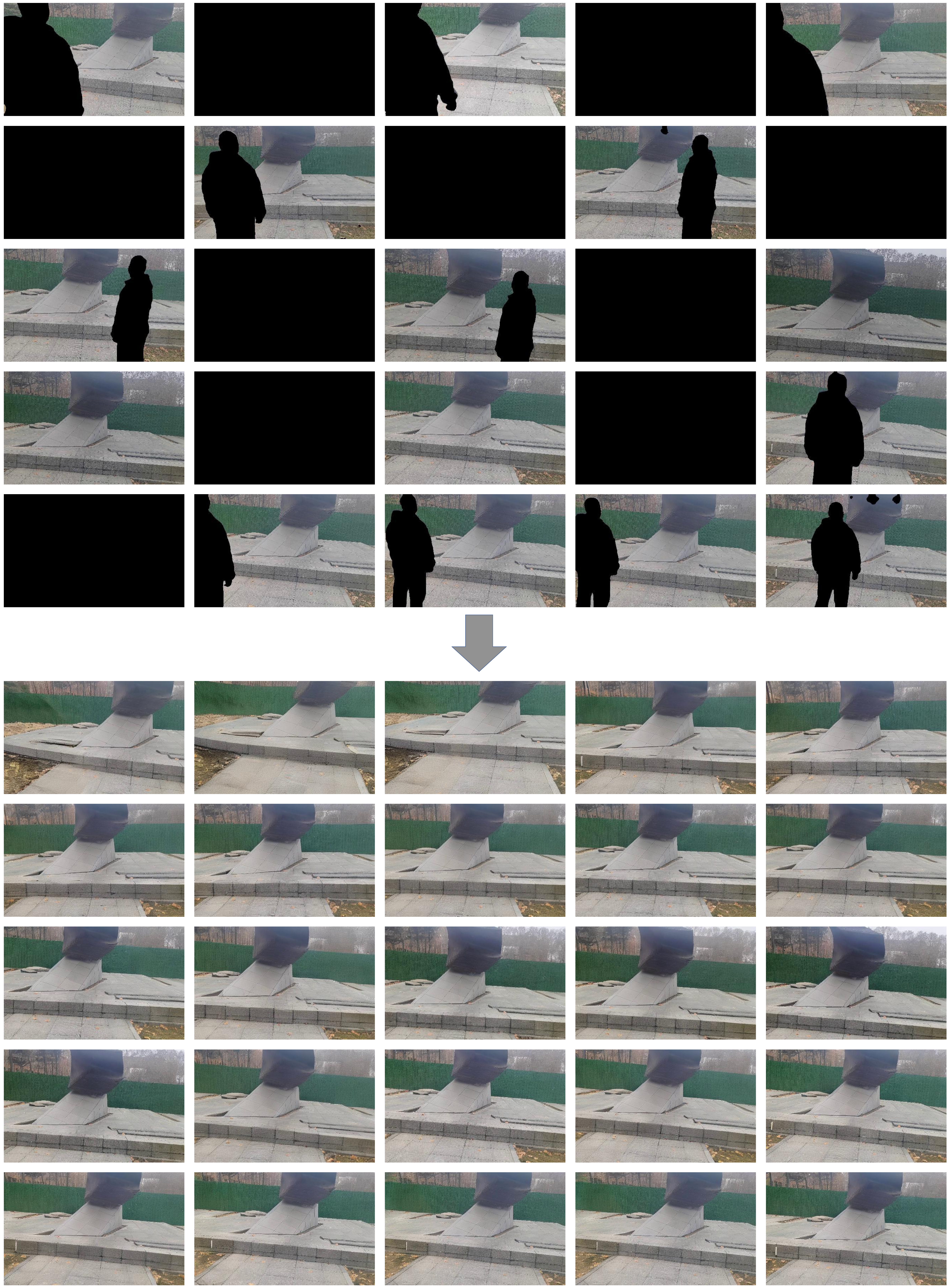}
    \caption{Visualization of how UniVerse turns a initial video into restored video.}
    \label{fig:app_video_sculp}
\end{figure*}

Since UniVerse utilizes a VDM to turn initial videos into restored videos, we present several examples in Figs.~\ref{fig:app_video_horn}, \ref{fig:app_video_fern}, and \ref{fig:app_video_sculp}, demonstrating how UniVerse leverages the video prior to transform multi-view images into a consistent video. In these figures, the top row shows the initial video frames, while the bottom row displays the corresponding restored video frames. The frames are arranged from left to right in sequential order, with the first row showing frames 1-5, the second row showing frames 6-10, and so on.

\end{document}